# Multi-Scale Incremental Modeling for Enhanced Human Motion Prediction in Human-Robot Collaboration

Juncheng Zou [a, *]

[a] School of Electronic Information and Electrical Engineering, Huizhou University, Huizhou 516000, PR China

## Abstract

Accurate human motion prediction is crucial for safe human-robot collaboration but remains challenging due to the complexity of modeling intricate and variable human movements. This paper presents Parallel Multi-scale Incremental Prediction (PMS), a novel framework that explicitly models incremental motion across multiple spatio-temporal scales to capture subtle joint evolutions and global trajectory shifts. PMS encodes these multi-scale increments using parallel sequence branches, enabling iterative refinement of predictions. A multi-stage training procedure with a full-timeline loss integrates temporal context. Extensive experiments on four datasets demonstrate substantial improvements in continuity, biomechanical consistency, and long-term forecast stability by modeling inter-frame increments. PMS achieves state-of-the-art performance, increasing prediction accuracy by 16.3%-64.2% over previous methods. The proposed multi-scale incremental approach provides a powerful technique for advancing human motion prediction capabilities critical for seamless human-robot interaction.

## 1 Introduction

Human-robot collaboration presents a transformative prospect applicable to diverse scenarios such as manufacturing, surgery, disaster relief, and home care. However, effective human motion forecasting over long time horizons remains an open challenge. Illustratively, envision an automated production set-up wherein robots operate alongside humans. In such instances, the necessity for robots to precisely anticipate human motions and intentions for the dual purpose of efficient operation and safety are paramount. A flawed estimation of a human's planned action could culminate in accidents, productivity interruptions, or efficiency drop-offs. This necessitates an accurate prognostication of human movements, which forms the bedrock of our research. Despite the evident advantages of such interactions, they frequently confront intractable hurdles due to the inherent complexity, variability, and unpredictability associated with human movement.

[*] Corresponding author. E-mail addresses: zoujc752@163.com (J. Zou).

For instance, cluttered environments prompt swift amendments in human navigation trajectories, thereby confounding robot planning algorithms. This leads to discontinuities, error propagation, and unrealistic motion trajectories that fail to capture the nuanced dependencies and spatial-temporal dynamics critical for natural human movement.

Despite intense research, existing human motion prediction techniques still struggle with effectively handling the complex biomechanics, temporal dynamics, and multi-step uncertainty propagation for longer-term forecasts. For instance, while data-driven deep learning models show promise, they fail to effectively address multi-step uncertainty propagation and lack an inductive bias capable of modeling physical motion over extended timeframes. RNN architecture often registers discontinuities due to error accumulation across iterations. Additionally, most existing techniques emphasize predicting absolute pose targets rather than incremental transitions between frames. However, this fails to capture the continuous spatial-temporal dynamics critical for realistic motion modeling [12]. While mining inter-frame incremental information has been identified as a crucial technique to augment HMP performance, with potential for further enhancements by combined utilization of both incremental information [40] and global context, existing methods have shown limitations. Methods leveraging incremental modeling for HMP through extraction of inter-frame differences [1-3] or parallel decoding [4] have demonstrated shortcomings in competently capturing multi-scale dynamics or handling uncertainty over an extended prediction horizon.

To address this, our goal is to develop a new parallel prediction framework that can effectively model both local and global human motion dynamics for improved safety and productivity in long-term human-robot collaboration scenarios. Specifically, we propose PMS - a parallel prediction framework with three key innovations:

1) A multiscale incremental computation module that captures multiscale motion increments and dynamics at different time intervals. This enables the modeling of both short-term and long-term dependencies in motion patterns. Importantly, by computes velocity and acceleration differences across multiple timescales, our approach can effectively mine informative prior patterns from historical motion data at different granularities. This data-driven extraction of multiscale incremental dynamics represents a key novelty, allowing the model to learn latent representations that capture subtle details of motion history dependencies. The learned motion priors help enhance future forecasting quality.

2) A parallel prediction mechanism with an iterative graph adjustment layer to integrate the speed and acceleration differences calculated at different time intervals. This allows encoding both local and global motion dynamics. The module prediction branch then predicts future velocity and acceleration increments.

3) A full-time loss has local short-term and global long-term motion dependence. This jointly optimizes the prediction accuracy over the past, present, and future time horizons. This incorporates long-range context information. Together, these components provide a flexible, accurate, and simple approach for human motion forecasting, validated through extensive experiments showing significant improvements in handling biomechanics and uncertainty during long-term prediction. The multi-scale incremental extraction and parallel iterative adjustment make PMS uniquely capable over longer horizons.

In conclusion, PMS proposes a new parallel prediction framework to specifically address long-standing limitations in long-term uncertainty propagation, a key gap in human motion forecasting. We demonstrate PMS achieves significant performance gains on diverse benchmark tasks. The proposed techniques provide a robust solution for real-world motion prediction challenges involving human-robot interaction and collaboration over extended future periods.

The remainder of this paper is organized as follows. Section 2 reviews related work on human motion prediction, with a focus on incremental modeling approaches. Section 3 then describes our proposed PMS prediction framework, including the overall methodology, network architecture, and key technical innovations. Next, Section 4 presents our experimental results, covering the data setup, implementation details, performance benchmarking against existing state-of-the-art methods, and ablation studies analyzing the impact of different model components. Section 5 discusses the results, limitations, and potential societal impacts of our approach in detail. Finally, Section 6 concludes with a summary of key findings and directions for future work to build on this research.

# 2 Related Work

Human motion prediction has attracted substantial research attention, with various modeling paradigms explored. In this section, we provide a comprehensive review of the most relevant existing techniques, organized into three main categories: deep learning-based methods, inter-frame variation modeling, and motion feature-based approaches.

## 2.1 Deep Learning-based Methods

Deep learning techniques have extensively contributed to progress in human motion prediction with their adaptive capacity of capturing spatial-temporal dependencies in data. However, the recognition of intricate dependencies remains a challenge to be addressed. Convolutional Neural Networks (CNNs) [36, 37], for instance, show promising results in extracting local motion features and encoding spatial relationships [5-7]. They may find it challenging, however, to handle the complex multi-scale feature fusion and modeling long-term dependencies since they primarily operate on fixed-sized local receptive fields, thus capturing global contextual information might be challenging [40, 53].

On the other hand, Graph Convolutional Networks (GCNs) have demonstrated considerable adeptness at identifying intricate spatial relationships between body joints and effectively integrating multi-scale features [8-15]. Nonetheless, it's worth considering that these networks may encounter hurdles when faced with spatial-temporal data irregularities such as inconsistency in motion sequence lengths or missing data points. These challenges arise from GCN's inherent assumptions about consistently structured graphs and fixed input sizes [41, 54]. This necessitates intense hyperparameter tuning, including selecting the right graph convolution operators, neighborhood definitions, and regularization strategies, to derive the best possible performance from GCNs [55].

Continuing this trend, transformer-based models [16-19] have recently attracted a lot of attention due to their proficiency in capturing long-range dependencies in time series data. It is, however, worth mentioning that such advanced models often require careful regularization in order to prevent overfitting, especially when training data is sparse [42, 56].

While deep learning techniques have significantly advanced the field, generating realistic and generalizable motion predictions solely based on data-driven learning, without explicitly modeling motion dynamics and features, remains a challenge. Purely data-driven approaches may struggle to capture the inherent physical constraints and biomechanical principles governing human motion, leading to predictions that violate natural motion patterns or plausibility [43, 44].

## 2.2 Inter-frame Variation Modeling

Explicitly modeling the incremental transformations between frames has emerged as key for ensuring temporally coherent predictions over long horizons. Methods have aimed to strike a balance between precision and computational complexity. Most existing approaches focus on predicting absolute human joint positions, neglecting the importance of modeling inter-frame motion transformations. However, incremental transformation prediction has been shown to be less susceptible to accuracy degradation due to incorrect actions and is vital for achieving continuous long-term prediction [45, 46].

Several methods have been proposed to explicitly model inter-frame variations, such as mutation networks [23], geodesic loss functions [24], spatio-temporal branch networks [1], residual blocks [2], and differential generation combined with attention mechanisms [3]. These approaches aim to capture local and global motion information, improve temporal connectivity, and enhance long-term prediction accuracy.

However, even these promising methods come with some limitations. For instance, geodesic loss-based methods can be computationally intensive due to the calculation of geodesic distances on non-Euclidean manifolds [24]. Furthermore, some commonly used techniques like attention mechanisms might incur high computational costs when dealing with long sequences or high-dimensional data [3], and the performance of residual blocks heavily depends on careful hyperparameter tuning [2].

## 2.3 Motion Feature-based Approaches

Incorporating explicit motion domain knowledge and biomechanical constraints have shown promise in improving prediction plausibility. However, balancing model interpretability, accuracy and efficiency remains an open challenge.

Several techniques for incorporating motion features and prior knowledge into prediction models have been explored. Some of these features can be derived from physical laws [27-30], motion phases and subsequent elements [31-33], or specific techniques designed to reduce prediction errors [34, 35]. Notably, physics-based methods apply principles of human dynamics and biomechanics to add physical constraints and plausibility in motion predictions [27-30]. On the downside, these methods might be computationally demanding due to the simulation of complex physical interactions and might fail to capture both physical and cognitive elements of human motion .

Different methods have been proposed including the decomposition of motion into phases or components such as trajectories and poses [31, 32], and the utilization of specialized architectures like MotionGRU [33] to model transient changes and trends. However, these methods too, bring their own complexities due to the intricate interactions in human motion.

Other techniques aimed at improving long-term prediction accuracy while reducing computational complexity have also been introduced. These involve parallel decoding [4], key pose prediction [34], and dynamic correlation modelling [35]. While these methods contribute to the field, they also tend to introduce additional approximations or constraints that could limit their generalizability or fail to capture the subtleties in human motion.

Despite researchers making significant strides in human motion prediction, challenges persist:

1) The integration of spatial and temporal information across various scales remains difficult for current methods [47, 48]. By leveraging multi-scale incremental modeling, our proposed PMS approach aims to capture both fine-grained local dynamics and global trajectories.

2) Explicitly modeling inter-frame transformations remains crucial for avoiding discontinuities and artifacts [49, 50]. Our incremental prediction framework focuses precisely on encoding motion variations across frames and scales to ensure continuity.

3) Propagating errors can lead to unrealistic long-term forecasts [51, 52]. Our parallel prediction mechanism generates multiple candidate future sequences to improve robustness and accuracy over longer horizons.

In summary, through multi-scale incremental modeling and parallelization, the proposed PMS approach addresses key gaps in existing research by improving integration of spatial-temporal information, ensuring inter-frame coherence, and enhancing robustness for long-term forecasting. As validated through comprehensive benchmark evaluation, PMS represents a novel framework advancing state-of-the-art human motion prediction.

# 3 Method

## 3.1 Methodology Overview

Problem Definition: The research problem tackled in this paper is human motion prediction, specifically, forecasting a sequence of $L$ future pose frames based on $K$ observed frames. Here, the total sequence length is $T = K + L$. The input comprises $K$ historic observations $X_{1:K} = \{X_1, ..., X_K\}$ of dimension $(K, N, 3)$, with each $X_t \in R^{N \times D}$ representing the human pose at time t as $N$ joints in D spatial dimensions.

Specifically, the i-th pose frame $P_i$ contains N joint coordinate tuples $P_i = \{J_{i,1}, J_{i,2}, ..., J_{i,N}\}$, where each tuple $J_{i,k} = \{x_{i,k}, y_{i,k}, z_{i,k}\}$ gives the 3D coordinates of joint k.

The objective is to predict future poses $\widehat{X}_{K+1:K+L} = \{\widehat{X}_{K+1}, ..., \widehat{X}_{K+L}\}$ that minimize the error between the predictions and ground truth poses $X_{K+1:K+L} = \{X_{K+1}, ..., X_{K+L}\}$. The predictions are of dimension $(L, N, 3)$.

## 3. 2 PMS network

Existing methods for human motion prediction have limitations in modeling long-term dependencies. Prior works have explored pose-based prediction [20], multi-stage frameworks with graph convolutions [40], and motion attention mechanisms [47]. However, they lack sufficient capacity to capture joint dynamics across long time horizons.

To address these challenges, we propose a parallel prediction approach with a multi-scale incremental modeling for improved long-term forecasting. The proposed model (see Figure 1) comprises five sets of normalized joint position coordinates from a human skeleton model as input. Specifically, our contributions include:

1) A multiscale computation module calculating the velocity and acceleration differences from the input to capture multi-scale motion dynamics, shown in blue and orange in Figure 1, respectively. This module advances frame-level [47] or subsequence-level modeling by representing joint dynamics across multiple time intervals. By modeling these incremental changes in joint velocity and acceleration across different timescales, our approach can effectively mine informative prior patterns from historical motion data at various granularities. The rationale behind this is that human motion comprises both short-term and long-term dynamics, and capturing these incremental changes at multiple timescales allows the model to better represent and learn these intricate patterns.

2) A fusion module, which is the first column in Figure 1, then synthesizes the velocity and acceleration differences into multi-scale features. A dedicated differential learning module (shown in green in Figure 1) predicts velocity increments and acceleration changes using these fused features as input. The velocity prediction is consequently corrected by the acceleration prediction before the final adjustment of the motion prediction (shown in yellow in Figure 1). The fusion module is updated by an iterative adjustment mechanism and can represent global dynamics in parallel across time scales, in contrast to sequential prediction [40]. This parallel and iterative approach aims to capture the complex interplay between velocity and acceleration dynamics, enabling more accurate and robust long-term motion modeling.

3) An optimized full-time loss function incorporating long-range context across horizons, advancing limited context in prior losses [20, 40, 47]. By considering past, present, and future time steps in the loss function, the model can better learn and account for both short-term and long-term dependencies, leading to improved predictions across the entire motion sequence.

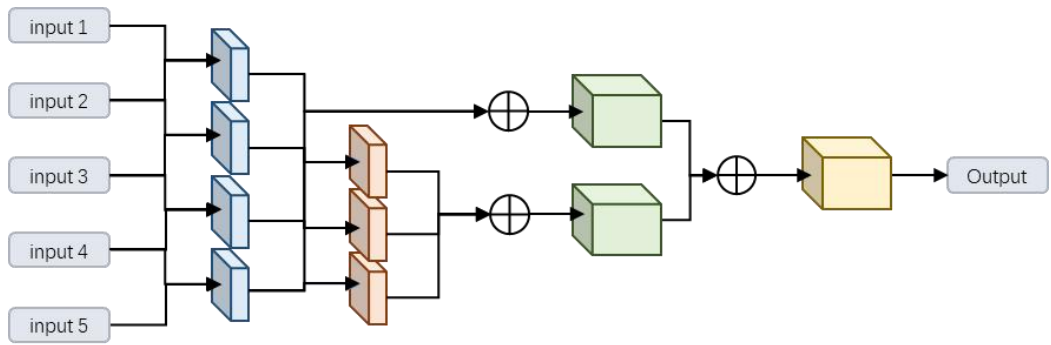

**Figure 1**. PMS framework. This network uses normalized joint position coordinates as input (from a human skeleton model) to capture and process motion dynamics. It employs a series of computational

modules (velocity and acceleration differences calculation, fusion, and differential learning) and an iterative adjustment mechanism to generate accurate predictions of human poses over time.

Together, the proposed modularized parallelism, multi-level incremental modeling, and full temporal awareness facilitate advanced handling of biomechanics and uncertainty over longer durations. Our approach provides more flexible, accurate and robust motion forecasts, advancing state-of-the-art in long-term human motion predictability.

**1. Normalization.** To enhance the capability of our network to learn motion patterns, each motion sequence $X_{1:K} = \{X_1, ..., X_K\}$ for a specific action $X_r$ is subjected to a normalization process. The normalization process involves several steps:

i) We first determine the maximum values $\{x_{max}, y_{max}, z_{max}\}$ and minimum values $\{x_{min}, y_{min}, z_{min}\}$ along the x, y, and z dimensions respectively, of action $X_r$.

ii) We then centered the interval by subtracting the mean of the maximum and minimum values from the original coordinates.

$$\begin{aligned} x_r' &= x_r - (x_{max} + x_{min})/2 \\ y_r' &= y_r - (y_{max} + y_{min})/2 \\ z_r' &= z_r - (z_{max} + z_{min})/2 \end{aligned} \tag{1}$$

Where $x_r$, $y_r$, and $z_r$ are the original coordinates.

iii) Each centered element $X_r'$ is then scaled to the range [-1,1] by division with half of the respective range yielding normalized coordinates:

$$\begin{aligned} x_r'' &= \frac{x_r'}{|x_{max} - x_{min}|/2} \\ y_r'' &= \frac{y_r'}{|y_{max} - y_{min}|/2} \\ z_r'' &= \frac{z_r'}{|z_{max} - z_{min}|/2} \end{aligned} \tag{2}$$

$\{x_r'', y_r'', z_r''\}$ denotes the normalized coordinates of $X_r'$.

**2. Multi-scale incremental fusion parallel prediction module**

This module aims to capture both short-term and long-term motion patterns by modeling incremental changes in joint velocity and acceleration across multiple timescales.

**Module 1: Incremental computation.** The velocity and acceleration differences are computed over three different time intervals: $\delta$ = 10, 5, and 2 frames (Equation 3). This allows the model to represent variations in short-term and long-term dependencies of motion patterns. Capturing finer-grained incremental motions helps retain precise pose details over time rather than mean pose.

$$\begin{aligned} &\{X_{r:r+10}, X_{r+10:r+20}, X_{r+20:r+30}, X_{r+30:r+40}, X_{r+40:r+50}\} \\ &\{X_{r:r+5}, X_{r+5:r+10}, X_{r+10:r+15}, X_{r+15:r+20}, X_{r+20:r+25}\} \\ &\{X_{r:r+2}, X_{r+2:r+4}, X_{r+4:r+6}, X_{r+6:r+8}, X_{r+8:r+10}\} \end{aligned} \tag{3}$$

The speed difference is calculated at three time intervals:

$$\Delta X_k^\delta = X_{r+k*\delta:r+(k+1)*\delta} - X_{r+(k-1)*\delta:r+k*\delta} \tag{4}$$

Here, $\Delta X_k^\delta$ denotes the k-th velocity difference, $r$ denotes the starting position of the interval calculation, $k$ denotes the number of intervals, and $\delta$ denotes the time interval (10, 5, 2).

The acceleration difference is calculated from the velocity difference:

$$\Delta\Delta X_k^\delta = \Delta X_{k+1}^\delta - \Delta X_k^\delta \tag{5}$$

Let $\Delta\Delta X_k^\delta$ denote the k-th acceleration difference.

The velocity and acceleration differences are then combined using multi-scale weighted combinations (Equations (6) and (7)) to synthesize comprehensive velocity and acceleration features. This weighted combination allows the model to capture motion patterns at different timescales, with the weights determining the relative importance of each timescale. By integrating velocity and acceleration differences over multiple time scales, the model can capture both short-term and long-term motion patterns to improve modeling and prediction. Multi-scale weighted combination to synthesize velocity features:

$$X_v^\delta = \sum_{i=1}^{N1} \alpha_i^\delta \Delta X_i^\delta \qquad (6)$$

Among them, $X_v^\delta$ represents the speed difference with an interval of δ, $\alpha_i^\delta$ is the coefficient of the speed difference, $\sum_{i=1}^{N1} \alpha_i = 1$, the setting of the coefficient will significantly affect the learning and prediction of the network. $N1$ is the number of speed differences.

Multi-scale weighted combination comprehensive acceleration feature:

$$X_a^\delta = \sum_{i=1}^{N2} \beta_i^\delta \Delta\Delta X_i^\delta \qquad (7)$$

The $X_a^\delta$ said for delta acceleration difference between integrated, $\beta_i^\delta$ the delta for acceleration differential coefficient, $\sum_{i=1}^{N2} \beta_i^\delta = 1$. $N2$ is the number of acceleration differences. By integrating velocity and acceleration differences over multiple time scales, the model can capture both short-term and long-term motion patterns to improve modeling and prediction.

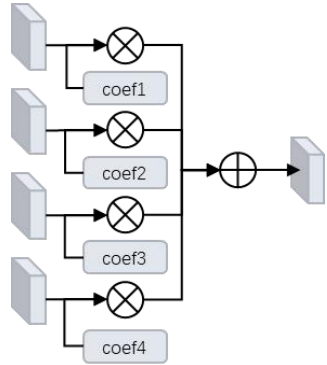

**Figure 2.** The incremental fusion module.

**Module 2: Incremental Learning Module.** This module consists of two submodules: Velocity Difference Incremental Learning and Acceleration Incremental Difference Learning. Their primary function is to predict future motion increments.

In the velocity difference learning process (Equations (8) to (11)), the synthesized velocity features are first processed by a fully connected layer, followed by an Long Short-Term Memory (LSTM) layer to capture temporal dependencies. The output of the LSTM is then passed through two additional fully connected layers with batch normalization and dropout for regularization. This submodule learns to predict the velocity increments based on the input velocity features. Speed difference learning:

$$_1X_v^\delta = \sum_{m=1}^{K} X_v^\delta W_1^\delta + b_1^\delta \qquad (8)$$

Here, $W_1^\delta$ and $b_1^\delta$ represent the network layer weights and offsets for $X_v^\delta$ velocity synthesis, respectively. The LSTM layer is used to capture temporal dependencies.

$$_2X_v^\delta = LSTM(\,_1X_v^\delta) \tag{9}$$

Fully connected layers, regularized using batch Normalization (BN) and dropout (DP):

$$_3X_v^\delta = \sigma\left(DP\left(BN\left(\sum_{m=1}^{K} {}_2X_v^\delta W_2^\delta + b_2^\delta\right)\right)\right) \tag{10}$$

Here, $BN$ represents batch normalization, $DP$ represents dropout, and $\sigma$ represents the activation function. $W_2^\delta$ and $b_2^\delta$ denote the network layer weights and offsets of $_2X_v^\delta$ velocity synthesis, respectively.

$$_4X_v^\delta = \sum_{m=1}^{K} {}_3X_v^\delta W_3^\delta + b_3^\delta \tag{11}$$

Similarly, in the acceleration difference learning process (Equations (12) to (15)), the synthesized acceleration features are processed through the same network architecture to predict the acceleration increments. Acceleration difference learning:

$$_1X_a^\delta = \sum_{m=1}^{K} X_a^\delta W_1^\delta + b_1^\delta \tag{12}$$

$$_2X_a^\delta = LSTM(\,_1X_a^\delta) \tag{13}$$

$$_3X_a^\delta = \sigma\left(DP\left(BN\left(\sum_{m=1}^{K} {}_2X_a^\delta W_2^\delta + b_2^\delta\right)\right)\right) \tag{14}$$

$$_4X_a^\delta = \sum_{m=1}^{K} {}_3X_a^\delta W_3^\delta + b_3^\delta \tag{15}$$

The predicted acceleration increment is then used to correct the predicted velocity increment (Equation (16)). This step accounts for the interplay between velocity and acceleration dynamics, aiming to improve the overall motion prediction accuracy. Correction of acceleration to velocity:

$$_4X_v^\delta = {}_4X_v^\delta + {}_4X_a^\delta \tag{16}$$

Finally, the corrected velocity increment is used to predict the future pose by applying an attenuation regulation mechanism (Equation (17)). The attenuation coefficients $\gamma_n$ allow the model to adjust the predicted pose based on the velocity dynamics, further enhancing the prediction robustness. Prediction of attenuation regulation based on velocity:

$$\widehat{X}_{r+k*\delta:r+(k+1)*\delta} = X_{r+k*\delta:r+(k+1)*\delta} - \sum_{n=1}^{L} \gamma_n {}_4^n X_v^\delta \tag{17}$$

**3. Comprehensive Time Loss Function.** During the motion optimization step, to enable the model to learn both short-term and long-term motion dependencies, we propose the comprehensive time loss function, $\mathcal{L}_a$. This function optimizes the motion by jointly considering a single loss $\mathcal{L}_c$, a past loss $\mathcal{L}_p$, and a future loss $\mathcal{L}_f$. This intricate configuration allows us to cater to both local and global dependencies more effectively. By directly comparing the predicted poses to the ground truth at each precise timeframe, our full temporal loss function helps retain precise pose details over the past, present, and future prediction horizons. This prevents averaging and blurring of poses over time.

$$\mathcal{L}_a = \mathcal{L}_p + \mathcal{L}_c + \mathcal{L}_f \tag{18}$$

$\mathcal{L}_p$ is the sum of the L1 loss between the predicted 3D joint positions $\widehat{X}$ and ground truth $X$ over $\Delta$ past frames, summed over $\Delta = \{2, 5, 10\}$ frames:

$$\mathcal{L}_p = \sum_{\Delta \in \{2,5,10\}} \sum_{t=T-\Delta}^{\Delta} \left| X_{t-\Delta:t} - \widehat{X}_{t-\Delta:t} \right| \tag{19}$$

Here, $T$ denotes the length of the unit sequence.

$\mathcal{L}_c$ is the L1 loss between predicted and ground truth pose at the current time $t$:

$$\mathcal{L}_c = \sum_{t=0}^{T} \left| X_t - \widehat{X}_t \right| \tag{20}$$

$\mathcal{L}_f$ is the sum of L1 loss between predicted and ground truth pose $X$, over $\Delta$ future frames, summed over $\Delta = \{20, 30\}$ frames:

$$\mathcal{L}_f = \sum_{\Delta \in \{20, 30\}} \sum_{t=\Delta}^{T-\Delta} \left| X_{t:t+\Delta} - \widehat{X}_{t:t+\Delta} \right| \tag{21}$$

# 4 Experiment

In this study, we aim to address the following research questions:

RQ1: How does the proposed Parallel Multi-scale Incremental Prediction (PMS) method compare with existing state-of-the-art human motion prediction methods, and what is its overall performance?

RQ2: What insights can be gleaned from the experimental results regarding the contribution of the different components of PMS to its performance, and what are its potential limitations?

To answer these research questions, we conducted a comprehensive set of experiments on multiple benchmark datasets. The experimental setup and implementation details are presented in Sections 4.1 and 4.2, respectively. Section 4.3 addresses RQ1 by comparing the performance of PMS with various baseline methods, while Section 4.4 provides insights into RQ2 by analyzing the contribution of different components and discussing the limitations of the proposed approach. Statistical analyses, including measures of central tendency and variability, are provided to support the claims and comparisons made.

## 4.1 Data setup

1) **Human3.6M[48]:** This dataset comprises 3.6 million motion sequences spanning 15 action

categories performed by seven professional actors. Each pose consists of 32 joints, of which 28 are used by eliminating hand and foot joints that are either zero or repeated. Each data sequence is down-sampled to 25 fps. The training setup adheres to the methodology of previous studies [49-50], using Subject1, 6, 7, 8, 9 for training data, Subject5 for testing, and Subject11 for validation. We segmented each action into smaller sequences for training. In long-term forecasting, our model builds upon short-term forecasting techniques, applying our model in an autoregressive manner. The model inputs 50 frames and outputs 10 frames, which are then combined with the predicted 10 frames and the latest 40 frames of the history frames to form a new input. The process continues with 50 frames of input and 25 frames of output.

**2) CMU Mocap[1]:** To validate the generalizability of our Parallel Multi-scale Incremental Prediction (PMS), we conducted experiments on the CMU Mocap dataset. This dataset contains five classes of actions, each with several subclasses. Following [51], we selected eight actions for detailed analysis: basketball, basketball signal, directing traffic, jumping, running, football, walking, and washing windows. As previous studies did not clearly delineate the proportions for training, validation, and testing sets, we adopted a 70%, 15%, and 15% split, resulting in 174 training data, 38 validation data, and 36 testing data. The prediction settings are the same as in Human3.6M. The sampling frequency for this dataset is 120Hz.

**3) 3DPW[52]:** We also conducted experiments on the 3DPW dataset to further validate the generalizability of our proposed PMS. The 3DPW dataset, a challenging collection of 51,000 human motion poses from indoor and outdoor scenes, adheres to the official train/validation/test set split. The training settings are identical to those in Human3.6M. The dataset has a sampling frequency of 60 Hz, with each data sample containing 72 positions.

**4)AMASS-BMLrub:** The AMASS-BMLRUB dataset, part of the larger AMASS dataset [53], includes 111 individuals (50 males and 61 females) performing a total of 20 actions, amounting to 3061 actions over 522.69 minutes. Each sequence comprises between 200 and 2500 humanoid objects (frames), covering a total of 6980 vertices. Most previous researchers used AMASS-BMLRUB for testing, while other AMASS datasets served as training and validation sets. Given the extensive size of the AMASS dataset, training is a lengthy process. As our study aims to validate the model's generalization performance, we used only the ACCAD[2] dataset from AMASS for training, while AMASS-BMLRUB served as the test set. The ACCAD dataset includes 20 individuals, 252 actions, and spans 26.74 minutes. The sampling frequency for this dataset is 120Hz, with a length of 156 for individual data samples.

## 4.2 Implementation details

Implementation Details: All experiments were executed on an NVIDIA RTX 3090 24G using the PaddlePaddle framework [54]. All models were trained using the Adam optimizer [55] with an initial learning rate of 5e-3. For the Human3.6M dataset, we will first train each class of actions 10 times, then train 10 times by combining the single loss and the total loss of the accumulated 5 losses, and finally reduce the learning rate to 1e-3 for another 10 times. The training process also included ten iterations combining the long-term prediction loss. The memory consumption during training was approximately 2.6GB. For the CMU dataset, the PMS underwent ten training iterations. Similarly, for the 3DPW dataset, the PMS was trained ten times with the learning rate set to 1e-3. For the AMASS-BMLRUB dataset, the PMS was trained ten times using the ACCAD dataset. Across all four datasets utilized in this study, the input length was set to 50, the output length for short-term prediction was 10, and the output length for long-term prediction was 25, employing an autoregressive prediction approach. The Mean Per Joint Position Error (MPJPE) was employed as the loss function, which is a widely used metric in human motion prediction tasks. It measures the Euclidean distance between the predicted and ground truth joint positions, averaged over all joints and frames. Specifically, for a given sequence of T frames and J joints, the MPJPE is calculated as:

$$MPJPE = \left(\frac{1}{T*J}\right) * \Sigma_t \Sigma_j \left\| p_{tj} - p'_{tj} \right\| \tag{22}$$

where $p_{tj}$ and $p'_{tj}$ represent the ground truth and predicted 3D positions of joint $j$ at time $t$, respectively, and $\|\cdot\|$ denotes the Euclidean distance.

---

1  http://mocap.cs.cmu.edu/
2  https://accad.osu.edu/research/motion-lab/system-data

## 4.3 Comparison with existing methods (RQ1)

1) Baseline

To comprehensively evaluate the performance of the proposed PMS method, we compared it with 14 state-of-the-art baseline models spanning various architectures, including Graph Convolutional Networks (GCNs) [8, 36, 38, 42, 45, 46], fully connected layers [37, 43], Convolutional Neural Networks (CNNs) [6], Recurrent Neural Networks (RNNs) [39], Generative Adversarial Networks (GANs) [44], attention-based models [47], and stage-based models [40, 41].

The selection of these baseline models was based on their relevance to the problem domain and their reported state-of-the-art performance in human motion prediction tasks. By comparing against a diverse set of architectures and approaches, we aimed to provide a comprehensive evaluation of the proposed PMS method's performance relative to existing techniques.

The results of the comparison with existing methods are presented and analyzed for each dataset in the following subsections.

2) Results

H3.6M: The Mean Per Joint Position Error (MPJPE) comparison of our model's results on the Human3.6M dataset for short-term and long-term predictions is presented in Tables 1 and 2. Figures 3 and 4 illustrate the comparison of the mean MPJPE values for each action class.

Upon analyzing Table 1, it is evident that in the short-term prediction of the Human3.6M dataset, our PMS model falls short of the Motionmixer and Rele-GCN models in two time intervals of the "Walking" action. This is primarily due to insufficient prediction at 320ms and 400ms. Figure 3 demonstrates that the Motionmixer, Rele-GCN, and our PMS models outperform other methods in the short-term prediction of various types of actions. Except for the "Walking" action, the PMS model significantly surpasses other methods.

The PMS model, an MLP-based approach, outperforms other methods for all other actions, particularly for "Sitting down" and "Taking Photo" actions. For the "Sitting down" action, it is 64.2%, 57.4%, 49.0%, and 44.4% lower than the second-best method, Motionmixer, across four prediction periods. For the "Taking Photo" action, it is 63.6%, 54.8%, 43.9%, and 38.7% lower than Motionmixer across four prediction cycles. From the "Average" analysis, it is 70.0%, 47.0%, 29.4%, and 22.0% lower than Motionmixer across four time intervals.

These results indicate that the proposed PMS method significantly improves short-term prediction, and the speed incremental learning mode of PMS enhances performance. Figure 3 also shows that the MPJPE of the prediction method fluctuates depending on the actions, and mitigating such fluctuations is crucial for improving the generalization ability of the prediction method.

The mean MPJPE of each action can be calculated according to the MPJPE of each method in all actions, and the action variance of each method can be calculated according to the mean. Finally, the influence degree of the action on the prediction of the method can be judged according to the method. Upon computation, it was found that the action variance of the PMS method was 3.8. This value surpasses the second-best action variance of 6.0 from the Motionmixer method and the third-best action variance of 6.6 from the Rele-GCN method. These findings indicate that the predictive performance of the PMS method is less susceptible to variations in actions.

**Table 1.** Comparison of short term prediction results (in MPJPE) on the Human3.6M dataset.

| Action | Walking | | | | Eating | | | | Smoking | | | | Discussion | | | |
|---|---|---|---|---|---|---|---|---|---|---|---|---|---|---|---|---|
| Time(ms) | 80 | 160 | 320 | 400 | 80 | 160 | 320 | 400 | 80 | 160 | 320 | 400 | 80 | 160 | 320 | 400 |
| GA-MIN[36] | 7.5 | 13.5 | 28.2 | 30.6 | 5.8 | 12.5 | 23.5 | 33.8 | 6.2 | 12.1 | 24.2 | 24.2 | 8.2 | 18.6 | 30.5 | 46.3 |
| Motionmixer[37] | 7.3 | 12.9 | 23.5 | **28.6** | 4.3 | 8.3 | 16.9 | 20.9 | 4.7 | 8.8 | 17.3 | 21.4 | 6.4 | 13.1 | 28.6 | 35.5 |
| DSTD-GC[38] | 11.1 | 22.4 | 38.8 | 45.2 | 7.0 | 15.5 | 31.7 | 39.2 | 6.6 | 14.8 | 29.8 | 36.7 | 10.0 | 24.4 | 54.5 | 67.4 |
| DPnet[8] | 7.3 | 15.2 | 30.1 | 32.6 | 8.6 | 18.3 | 36.4 | 43.5 | 6.9 | 13.5 | 24.3 | 28.7 | 8.2 | 20.1 | 38.2 | 43.0 |
| Chopin et al.[39] | 12.0 | 21.1 | 35.6 | 42.4 | 9.0 | 15.9 | 29.1 | 35.0 | 7.9 | 14.3 | 25.2 | 30.4 | 11.9 | 22.7 | 44.8 | 54.6 |
| Ma et al.[40] | 10.2 | 19.8 | 34.5 | 40.3 | 7.0 | 15.1 | 30.6 | 38.1 | 6.6 | 14.1 | 28.2 | 34.7 | 10.0 | 23.8 | 53.6 | 66.7 |
| DeFeeNet[41] | 10.4 | 20.0 | 34.7 | 42.2 | 7.0 | 15.2 | 31.4 | 38.4 | 6.8 | 14.5 | 29.0 | 35.8 | 11.1 | 25.4 | 55.8 | 68.2 |
| Rele-GCN[42] | 6.5 | 12.4 | 20.7 | 28.7 | 5.2 | 8.8 | 18.4 | 21.8 | 5.2 | 9.6 | 18.4 | 22.8 | 6.7 | 16.4 | 28.6 | 35.2 |
| **PMS** | **3.0** | **8.0** | **24.2** | 35.3 | **2.1** | **5.2** | **14.3** | **20.0** | **2.0** | **4.8** | **12.8** | **17.8** | **2.6** | **6.9** | **19.1** | **26.3** |
| Action | Directions | | | | Greeting | | | | Phoning | | | | Posing | | | |
| Time(ms) | 80 | 160 | 320 | 400 | 80 | 160 | 320 | 400 | 80 | 160 | 320 | 400 | 80 | 160 | 320 | 400 |
| GA-MIN[36] | 6.8 | 15.3 | 42.1 | 50.2 | 12.8 | 26.3 | 61.8 | 75.8 | 8.3 | 17.8 | 37.9 | 44.8 | 7.8 | 19.3 | 43.4 | 56.0 |
| Motionmixer[37] | 4.4 | 9.7 | 22.5 | 29.2 | 8.8 | 17.7 | 36.9 | 46.2 | 5.6 | 10.7 | 21.9 | 27.8 | 6.0 | 13.1 | 30.2 | 40.1 |
| DSTD-GC[38] | 6.9 | 17.4 | 41.0 | 51.7 | 14.3 | 33.5 | 72.2 | 87.3 | 8.5 | 19.2 | 40.3 | 49.9 | 10.0 | 25.4 | 60.6 | 77.3 |
| DPnet[8] | 10.1 | 21.0 | 45.8 | 56.7 | 12.7 | 27.1 | 65.6 | 82.9 | 10.2 | 17.4 | 35.7 | 41.3 | 7.4 | 21.9 | 63.5 | 78.8 |
| Chopin et al.[39] | 11.1 | 20.9 | 38.8 | 47.0 | 19.6 | 35.1 | 64.0 | 78.2 | 11.7 | 19.4 | 34.9 | 42.3 | 13.7 | 25.9 | 50.0 | 61.1 |
| Ma et al.[40] | 7.2 | 17.6 | 40.9 | 51.5 | 15.2 | 34.1 | 71.6 | 87.1 | 8.3 | 18.3 | 38.7 | 48.4 | 10.7 | 25.7 | 60.0 | 76.6 |
| DeFeeNet[41] | 7.0 | 17.0 | 40.0 | 50.9 | 16.8 | 33.0 | 68.5 | 83.2 | 11.6 | 19.9 | 41.0 | 50.1 | 14.7 | 28.3 | 65.0 | 81.1 |
| Rele-GCN[42] | 6.2 | 11.9 | 27.1 | 32.9 | 10.8 | 18.9 | 37.8 | 44.1 | 7.9 | 11.2 | 24.1 | 26.8 | 5.3 | 15.2 | 33.8 | 44.2 |
| **PMS** | **1.8** | **5.1** | **14.7** | **20.5** | **3.6** | **9.6** | **26.5** | **36.3** | **2.4** | **6.0** | **16.1** | **22.3** | **2.0** | **5.9** | **17.3** | **24.2** |
| Action | Purchases | | | | Sitting | | | | Sitting Down | | | | Taking Photo | | | |
| Time(ms) | 80 | 160 | 320 | 400 | 80 | 160 | 320 | 400 | 80 | 160 | 320 | 400 | 80 | 160 | 320 | 400 |
| GA-MIN[36] | 12.4 | 28.5 | 60.0 | 72.9 | 8.1 | 18.5 | 41.9 | 53.2 | 14.5 | 25.5 | 56.3 | 70.3 | 8.3 | 19.6 | 38.2 | 49.0 |
| Motionmixer[37] | 8.4 | 16.9 | 34.1 | 42.7 | 6.5 | 11.8 | 23.6 | 29.8 | 10.9 | 18.8 | 35.1 | 42.6 | 5.5 | 10.4 | 22.1 | 27.9 |
| DSTD-GC[38] | 12.7 | 29.7 | 62.3 | 75.8 | 8.8 | 19.3 | 42.3 | 54.3 | 14.1 | 28.0 | 57.3 | 71.2 | 8.4 | 18.8 | 42.0 | 53.5 |
| DPnet[8] | 17.8 | 37.0 | 62.1 | 65.6 | 9.1 | 23.0 | 48.1 | 62.8 | 9.7 | 24.2 | 49.7 | 62.0 | 5.7 | 14.4 | 35.6 | 47.9 |
| Chopin et al.[39] | 14.2 | 26.5 | 48.3 | 58.1 | 10.4 | 17.9 | 33.1 | 40.7 | 15.8 | 28.2 | 52.9 | 64.5 | 11.7 | 19.4 | 34.9 | 42.3 |
| Ma et al.[40] | 12.5 | 28.7 | 60.1 | 73.3 | 8.8 | 19.2 | 42.4 | 53.8 | 13.9 | 27.9 | 57.4 | 71.5 | 8.4 | 18.9 | 42 | 53.3 |
| DeFeeNet[41] | 16.8 | 32.7 | 67.9 | 80.3 | 14.2 | 23.6 | 47.7 | 58.7 | 10.1 | 29.4 | 62.0 | 70.8 | 7.8 | 16.9 | 38.8 | 47.9 |
| Rele-GCN[42] | 10.5 | 18.5 | 39.7 | 45.5 | 6.5 | 13.7 | 28.5 | 36.9 | 8.8 | 22.7 | 39.6 | 45.9 | 5.1 | 12.5 | 26.9 | 33.5 |
| **PMS** | **3.4** | **9.0** | **23.3** | **31.0** | **2.6** | **5.8** | **13.7** | **18.4** | **3.9** | **8.0** | **17.9** | **23.7** | **2.0** | **4.7** | **12.4** | **17.1** |
| Action | Waiting | | | | Walking Dog | | | | Walking Together | | | | Average | | | |
| Time(ms) | 80 | 160 | 320 | 400 | 80 | 160 | 320 | 400 | 80 | 160 | 320 | 400 | 80 | 160 | 320 | 400 |
| GA-MIN[36] | 7.5 | 17.2 | 41.1 | 52.3 | 18.9 | 38.5 | 70.9 | 84.0 | 8.5 | 18.3 | 34.2 | 39.9 | 9.4 | 19.9 | 42.4 | 52.2 |
| Motionmixer[37] | 5.4 | 10.9 | 23.2 | 30.0 | 13.4 | 24.6 | 45.2 | 54.1 | 5.9 | 11.3 | 22.2 | 27.4 | 9.0 | 13.2 | 26.9 | 33.6 |
| DSTD-GC[38] | 8.7 | 20.2 | 44.3 | 55.3 | 19.6 | 41.8 | 77.6 | 90.2 | 9.1 | 19.8 | 36.3 | 42.7 | 10.4 | 23.3 | 48.8 | 59.8 |
| DPnet[8] | 8.4 | 20.5 | 53.6 | 69.1 | 25.7 | 51.8 | 94.9 | 112.3 | 8.3 | 18.8 | 35.6 | 44.8 | 10.3 | 22.9 | 47.9 | 58.1 |
| Chopin et al.[39] | 11.4 | 20.3 | 38.8 | 47.2 | 19.3 | 34.2 | 65.6 | 77.5 | 11.6 | 19.7 | 34.5 | 41.8 | 12.6 | 22.5 | 41.9 | 50.8 |
| Ma et al.[40] | 8.9 | 20.1 | 43.6 | 54.3 | 18.8 | 39.3 | 73.7 | 86.4 | 8.7 | 18.6 | 37.4 | 41.0 | 10.3 | 22.7 | 47.4 | 58.5 |
| DeFeeNet[41] | 9.6 | 19.8 | 42.3 | 53.6 | 17.6 | 41.1 | 72.7 | 84.9 | 8.8 | 19.0 | 36.1 | 41.8 | 11.4 | 23.7 | 48.9 | 59.2 |
| Rele-GCN[42] | 7.1 | 13.8 | 28.5 | 33.6 | 15.5 | 27.4 | 48.9 | 57.2 | - | - | - | - | - | - | - | - |
| **PMS** | **2.4** | **6.1** | **16.7** | **23.3** | **4.2** | **11.7** | **32.0** | **43.4** | **2.1** | **5.9** | **17.7** | **25.6** | **2.7** | **7.0** | **19.0** | **26.2** |

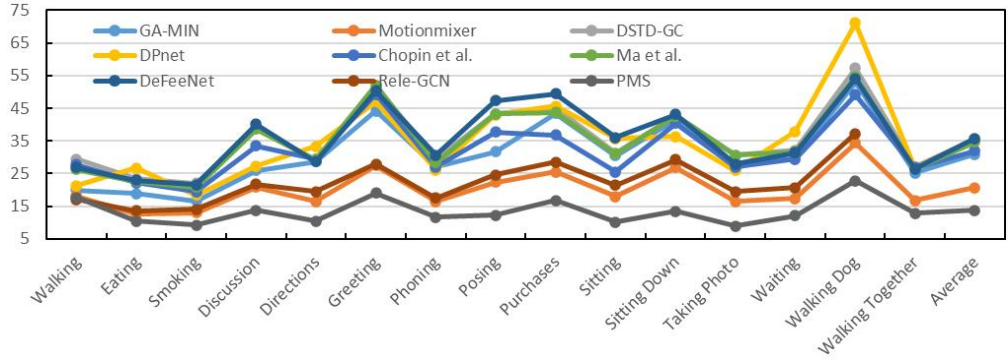

**Figure 3.** Comparison of short term prediction average results (in MPJPE) on the Human3.6M.

The long-term prediction comparison in Table 2 and Figure 4 shows that PMS outperforms other methods for 6 actions across the two "Average" time intervals. Specifically, PMS achieves lower MPJPE than other methods for 560ms predictions of 6 actions. For other actions, PMS is comparable or slightly weaker than other approaches. Overall, PMS achieves better performance on 20 out of 32 metrics, with 12 better predictions at 560ms and 8 at 1000ms. Notably, PMS significantly reduces MPJPE for "Sitting Down" and "Posing" compared to the next best method, Rele-GCN, by 40.9% and 36.9% at 560ms and by 16.3% and 20.2% at 1000ms, respectively. For the "Average" metrics, PMS lowers MPJPE by 37.8% and 17.0% versus the next best GA-MIN at the two time intervals. In the realm of long-term forecasting, Rele-GCN demonstrates the least average action variance, registering 13.3 and 17.5 respectively, while PMS follows closely with the second least average action variance, recording 11.0 and 30.9 respectively. In summary, PMS considerably diminishes the MPJPE at both 560ms and 1000ms intervals. Furthermore, it enhances the generalization of the long-term predictive action at the 560ms mark. However, it does not yield any improvement in the generalization of the predictive action at the 1000ms interval.

**Table 2.** Comparison of long term prediction results (in MPJPE) on the Human3.6M dataset.

| Action | Walking | | Eating | | Smoking | | Discussion | | Directions | | Greeting | | Phoning | | Posing | |
|---|---|---|---|---|---|---|---|---|---|---|---|---|---|---|---|---|
| Time(ms) | 560 | 1000 | 560 | 1000 | 560 | 1000 | 560 | 1000 | 560 | 1000 | 560 | 1000 | 560 | 1000 | 560 | 1000 |
| GA-MIN[1] | 35.5 | 42.8 | 47.3 | 65.2 | 30.6 | 46.5 | 60.3 | 106.3 | 68.1 | 100.0 | 95.3 | 121.5 | 63.0 | 101.5 | 83.2 | 150.2 |
| DSTD-GC[38] | 52.7 | 59.8 | 51.9 | 76.2 | 48.1 | 71.2 | 87.0 | 116.3 | 69.1 | 99.1 | 108.7 | 142.3 | 66.7 | 102.2 | 106.5 | 163.3 |
| DPnet[8] | 40.5 | 48.6 | 56.5 | 69.6 | 32.8 | 59.9 | 66.3 | 96.7 | 80.2 | 103.5 | 93.7 | 85.6 | 61.4 | 113.9 | 105.9 | 205.6 |
| Rele-GCN[42] | **31.9** | **37.6** | 27.4 | 48.3 | 28.1 | 44.2 | 50.1 | **74.3** | 42.8 | 64.5 | 58.2 | **85.5** | 39.7 | **64.2** | 62.8 | 98.2 |
| SIMLPE[43] | 46.8 | 55.7 | 49.6 | 74.5 | 47.2 | 69.3 | 85.7 | 116.3 | 73.1 | 106.7 | 99.8 | 137.5 | 66.3 | 103.3 | 103.4 | 168.7 |
| Trajectorycnn[6] | 37.9 | 46.4 | 59.2 | 71.5 | 32.7 | 58.7 | 75.4 | 103.0 | 84.7 | 104.2 | 91.4 | 84.3 | 62.3 | 113.5 | 111.6 | 210.9 |
| BiTGAN[44] | 49.8 | 60.5 | 48.5 | 73.0 | 48.4 | 70.0 | 85.8 | 116.4 | 73.3 | 106.3 | 101.1 | 136.4 | 67.3 | 103.2 | 107.1 | 171.0 |
| **PMS** | 62.5 | 158.4 | **33.0** | **74.2** | **29.4** | **66.6** | **41.9** | 81.1 | **33.0** | **64.0** | **57.1** | 107.8 | **36.7** | 80.2 | **39.6** | **78.4** |

| Action | Purchases | | Sitting | | Sitting Down | | Taking Photo | | Waiting | | Walking Dog | | Walking Together | | Average | |
|---|---|---|---|---|---|---|---|---|---|---|---|---|---|---|---|---|
| Time(ms) | 560 | 1000 | 560 | 1000 | 560 | 1000 | 560 | 1000 | 560 | 1000 | 560 | 1000 | 560 | 1000 | 560 | 1000 |
| GA-MIN[1] | 89.9 | 135.2 | 70.2 | 109.1 | 90.8 | 135.2 | 59.7 | 115.3 | 70.7 | 102.2 | 104.1 | 138.2 | 50.9 | 64.1 | 68.0 | 102.2 |
| DSTD-GC[38] | 97.5 | 137.8 | 74.9 | 117.8 | 96.1 | 147.3 | 74.5 | 117.9 | 73.2 | 105.7 | 109.8 | 147.7 | 50.5 | **61.2** | 77.8 | 111.0 |
| DPnet[8] | 94.2 | 123.2 | 72.5 | 106.9 | 84.6 | 131.0 | 74.0 | 83.2 | 96.7 | 167.7 | 136.7 | 174.9 | 59.8 | 78.1 | 77.0 | 109.9 |
| Rele-GCN[42] | 60.5 | 83.9 | 44.1 | 69.8 | 61.9 | 82.8 | 44.9 | 69.3 | 44.1 | **68.2** | 68.5 | **88.2** | - | - | - | - |
| SIMLPE[43] | 93.8 | 132.5 | 75.4 | 114.1 | 95.7 | 142.4 | 71.0 | 112.8 | 71.6 | 104.6 | 105.6 | 141.2 | 50.8 | 61.5 | 75.7 | 109.4 |
| Trajectorycnn[6] | 84.5 | 115.5 | 81.0 | 116.3 | 79.8 | 123.8 | 73.0 | 86.6 | 92.9 | 165.9 | 141.1 | 181.3 | 57.6 | 77.3 | 77.7 | 110.6 |
| BiTGAN[44] | 99.0 | 135.1 | 76.0 | 114.4 | 96.2 | 141.3 | 74.2 | 117.7 | 72.9 | 104.9 | 105.4 | 148.3 | 54.3 | 67.3 | 77.3 | 111.1 |
| **PMS** | **46.8** | **81.0** | **28.9** | **55.8** | **36.6** | **69.3** | **28.2** | **60.1** | **38.2** | 79.3 | **61.7** | 156.0 | **44.9** | 112.2 | **42.3** | **84.8** |

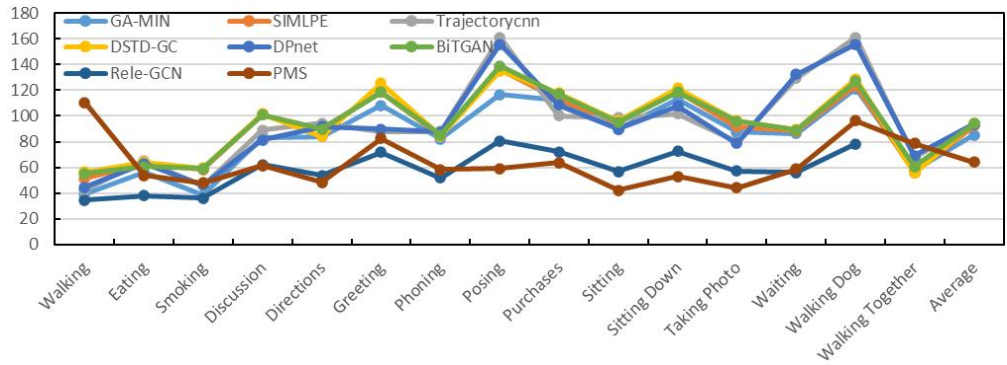

**Figure 4.** Comparison of long term prediction average results (in MPJPE) on the Human3.6M.

Upon examination of Figure 5, it becomes evident that there is a progressive increase in the Mean Per Joint Position Error (MPJPE) as the interval expands. The PMS method consistently surpasses other comparative methods across all time intervals. In particular, our PMS significantly outperforms the second best method GA-MIN at 320ms, 400ms and 560ms intervals. The MPJPE values for DSTD-GC and DPnet across various intervals exhibit a relative similarity.

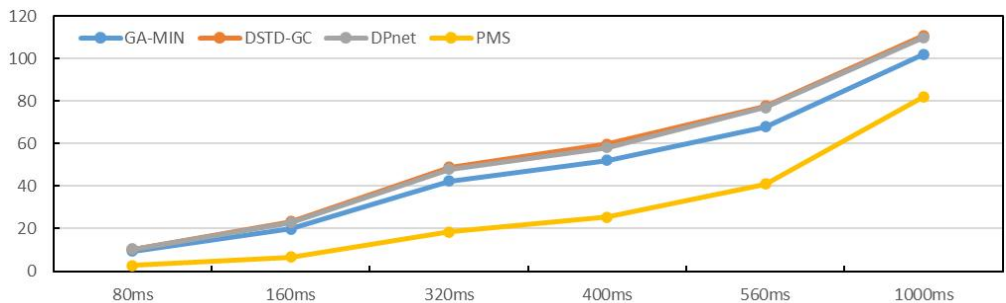

**Figure 5.** Comparison of prediction average results (in MPJPE) on the Human3.6M.

Figure 6 presents a comparative analysis between the actual trajectory and the Parallel Multi-scale Incremental Prediction (PMS) projected trajectory of a specific joint in human motion, as represented in the Human3.6M dataset. The figure illustrates that when the trajectory alterations are minimal and the trend of change remains relatively constant, the PMS's predicted trajectory closely aligns with the actual trajectory, as depicted in Figure 6(b), (c). However, when the trajectory trend undergoes substantial changes, a noticeable discrepancy emerges between the PMS prediction and the actual trajectory, as demonstrated in Figure 6(a), (d). This discrepancy becomes even more pronounced when the trajectory trend experiences significant alterations, as shown in Figure 6(e), (f). These observations suggest that the proposed PMS method is more adept at learning and predicting actions with gentle changes, while it struggles with actions involving large joint position changes and movements with conspicuous trend alterations. Therefore, further research is warranted to address this limitation in the PMS method.

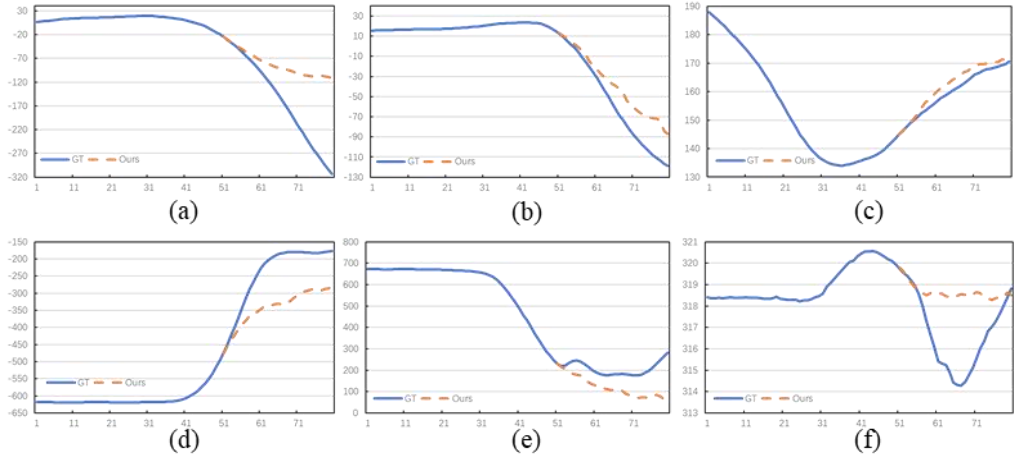

**Figure 6.** Comparison of predicted trajectories. (a) and (d) represent the action Walking. (b) and (e) represent the action Directions. (c) and (f) stand for the action Eating.

CMU. In the CMU study, a comparison of the "Average" across four intervals in Table 3 reveals that the MPJPE of the PMS method is 35.6%, 39.6%, 58.8%, 56.0% lower than that of the second-best method DSTD-GC, respectively. Notably, six types of movements are better learned by the PMS method, with "Jumping" showing significant improvement. The MPJPE reductions for "Jumping" are 57.6%, 58.6%, 70.3% and 68.6% across the four intervals, respectively, compared to the second-best method. The data suggests that the PMS method's performance enhancement becomes more pronounced as the time period lengthens. However, among the eight types of actions, "Basketball" and "Basketball signal" are less effectively learned by the proposed PMS method, with only two and one items respectively underperforming compared to other methods. This indicates that the PMS method's generalization still has potential for improvement. Furthermore, Chopin et al.'s method shows superior performance in "Basketball", achieving MPJPE optimality in two terms. The "Basketball signal" of Trajectorycnn is more optimal, with its MPJPE for one term outperforming several other methods.

**Table 3.** Comparison of prediction results on the CMU MOCAP dataset.

|  | Basketball | | | | Basketball signal | | | | Directing traffic | | | |
|---|---|---|---|---|---|---|---|---|---|---|---|---|
| Time(ms) | 80 | 160 | 320 | 400 | 80 | 160 | 320 | 400 | 80 | 160 | 320 | 400 |
| DSTD-GC[38] | 9.6 | 17.6 | 35.4 | 44.4 | 2.6 | 4.7 | 10.4 | 13.9 | 5.0 | 10.0 | 23.4 | 31.4 |
| Chopin et al.[39] | **9.1** | **16.6** | 34.7 | 44.5 | 3.3 | 5.9 | 11.5 | 14.7 | 19.6 | 31.3 | 54.8 | 66.1 |
| DPnet[8] | 10.7 | 17.8 | 38.4 | 49.5 | 2.6 | 4.4 | 10.0 | 13.4 | 5.9 | 11.8 | 26.6 | 33.5 |
| Trajectorycnn[6] | 11.1 | 19.7 | 43.6 | 56.8 | 1.8 | **3.5** | 9.1 | 13.0 | 5.5 | 10.9 | 23.7 | 31.3 |
| Mao et al.[45] | 10.7 | 20.0 | 42.2 | 53.9 | 3.0 | 6.0 | 14.5 | 19.4 | 5.6 | 11.8 | 28.7 | 38.9 |
| **PMS** | 11.5 | 18.8 | **25.2** | **34.5** | **1.7** | 3.6 | **5.3** | **7.1** | **4.3** | **8.4** | **12.1** | **16.0** |
|  | Jumping | | | | Running | | | | Soccer | | | |
| Time(ms) | 80 | 160 | 320 | 400 | 80 | 160 | 320 | 400 | 80 | 160 | 320 | 400 |
| DSTD-GC[38] | 12.8 | 26.1 | 54.6 | 68.5 | - | - | - | - | 10.3 | 19.0 | 36.8 | 45.7 |
| Chopin et al.[39] | 12.5 | 22.7 | 44.4 | 55.8 | 12.4 | 19.7 | 32.3 | 39.0 | 4.9 | 7.9 | 14.2 | 18.0 |
| DPnet[8] | 12.4 | 28.3 | 70.2 | 89.2 | 16.7 | 18.4 | 19.6 | 25.1 | 9.0 | 17.1 | 35.8 | 48.7 |
| Trajectorycnn[6] | 12.2 | 28.8 | 72.1 | 94.6 | 17.1 | 24.4 | 28.4 | 32.8 | 8.1 | 17.6 | 40.9 | 51.3 |
| Mao et al.[45] | 15.9 | 30.6 | 60.8 | 75.5 | 13.8 | 20.0 | 27.8 | 33.4 | 12.1 | 21.3 | 40.3 | 49.9 |
| **PMS** | **5.3** | **9.4** | **13.2** | **17.5** | **7.6** | **13.3** | **18.5** | **24.3** | **4.1** | **7.5** | **10.6** | **14.1** |
|  | Walking | | | | Wash window | | | | Average | | | |
| Time(ms) | 80 | 160 | 320 | 400 | 80 | 160 | 320 | 400 | 80 | 160 | 320 | 400 |
| DSTD-GC[38] | 6.3 | 10.4 | 16.1 | 18.6 | 4.8 | 9.5 | 22.0 | 29.0 | 7.3 | 13.9 | 28.4 | 35.9 |
| Chopin et al.[39] | 8.1 | 13.6 | 22.1 | 26.1 | 5.5 | 9.8 | 19.2 | 24.3 | 9.4 | 15.9 | 29.2 | 36.1 |
| DPnet[8] | 5.8 | 9.0 | 17.2 | 21.4 | 4.5 | 9.8 | 27.4 | 36.7 | 8.5 | 14.6 | 30.7 | 39.7 |
| Trajectorycnn[6] | 6.5 | 10.3 | 19.4 | 23.7 | 4.5 | 9.7 | 29.9 | 41.5 | 8.4 | 15.6 | 33.4 | 43.1 |
| Mao et al.[45] | 5.8 | 9.3 | 15.4 | 18.9 | 4.9 | 10.0 | 23.5 | 30.8 | 9.0 | 16.1 | 31.7 | 40.1 |
| **PMS** | **3.6** | **6.6** | **9.4** | **12.6** | **3.2** | **6.2** | **8.9** | **11.7** | **4.7** | **8.4** | **11.8** | **15.8** |

3DPW. On this dataset, Table 4 demonstrates that the Motion Att. + Post-fusion method surpasses other techniques, with the exception of the proposed method. When compared to the Motion Att. + Post-fusion method, the MPJPE of the proposed PMS method is reduced by 76.3%, 85.7%, 86.7%, 85.8%, 86.9%, 89.2% and 86.0% at six time intervals and the "Average", respectively. This suggests that predictive incremental learning can significantly enhance the prediction performance of our PMS method.

Table 4. Comparison of prediction results on the 3DPW dataset.

| Time(ms) | 80 | 160 | 320 | 400 | 560 | 1000 | Average |
|---|---|---|---|---|---|---|---|
| PhysMoP[56] | 0.7 | 3.6 | 15.1 | 19.8 | 30.3 | 70.9 | 23.4 |
| Motion Att. + Post-fusion[47] | 0.38 | 0.63 | 0.98 | 1.13 | 1.30 | 1.66 | 1.0 |
| PMS | 0.09 | 0.09 | 0.13 | 0.16 | 0.17 | 0.18 | 0.14 |

AMASS-BMLrub. The superiority of Motionmixer over the other 2 comparison methods is shown in Table 5. Compared with Motionmixer, our PMS reduces 79.2%, 63.5%, 55.5%, 89.7%, 92.0%, 6.7% and 15.3% in 7 indicators, respectively. It can be seen that the short-term and long-term prediction performance of PMS generalizes well.

Table 5. Comparison of prediction results on the AMASS-BMLrub dataset.

| Time(ms) | 80 | 160 | 320 | 400 | 560 | 1000 | Average |
|---|---|---|---|---|---|---|---|
| Motionmixer [37] | 6.6 | 10.3 | 18.0 | 21.9 | 28.8 | 41.6 | 21.2 |
| SIMLPE[43] | 10.8 | 19.6 | 34.3 | 40.5 | 50.5 | 65.7 | 36.9 |
| STSGCN[46] | 10.0 | 12.5 | 21.8 | 24.5 | 31.9 | 45.5 | 24.4 |
| PMS | 1.37 | 3.76 | 8.01 | 2.26 | 2.31 | 9.70 | 4.57 |

## 4.4 Effects of different compositions (RQ2)

To gain insights into the contribution of different components of the proposed PMS method and their impact on performance, we conducted two sets of ablation experiments comparing short-term and long-term predictions. The aim was to investigate the following hypotheses:

H1: The network architecture and loss function design significantly influence the prediction performance of PMS. This hypothesis is motivated by the understanding that the choice of network architecture and loss function can greatly impact the learning capabilities and optimization behavior of a neural network model.

H2: The integration of complementary velocity regulation layers with different time intervals contributes to the effectiveness of PMS in reducing the Mean Per Joint Position Error (MPJPE) for motion prediction. This hypothesis stems from the intuition that human motion involves complex temporal dynamics and capturing these dynamics at multiple time scales could enhance the model's ability to make accurate predictions.

The first set of experiments analyzed the effects of modifying the network settings and loss functions, using Baseline 1 as the reference configuration. Baseline 1 comprised a three-layer Long Short-Term Memory (LSTM) network with speed attenuation adjustment performed by sampling subtraction at intervals of 10, 5, and 2, with gradually decreasing attenuation coefficients. A fully connected layer was established before the speed incremental learning module, and four fully connected layers were set after this module, with a dropout rate of 0.4.

Five variations were compared against Baseline 1: (1) "FC w/ bias": Adding bias to the fully connected layers reduced the average MPJPE by 2.0% compared to Baseline 1. (2)"w/ bn+relu": Incorporating batch normalization and ReLU layers reduced the average MPJPE by 0.6%. (3)"w/ 5 times loss": Performing an extra update using the sum of 5 predicted loss values lowered the average MPJPE by 2.6%. (4) "w/ 2 5 10 times loss": Updating the network weights thrice using the aggregate loss values computed based on 2, 5, and 10 time periods resulted in an average increase of 4.7% in MPJPE compared to Baseline 1. (5) "w/ 800ms prediction loss": Updating the network weights by predicting the first 10 frames of 20 frames and adding the next 10 frames increased the average MPJPE by 3.2% compared to Baseline 1. The results of the first three variations support hypothesis H1, demonstrating the importance of network architecture and loss function design in improving prediction performance. The "w/ 2 5 10 times loss" and "w/ 800ms prediction loss" variants suggest that frequent updates to the network weights or directly updating weights twice through the prediction of 20 frames may introduce uncertainty and degrade performance.

Baseline 2 was constructed by enhancing the network training based on Baseline 1 and integrating the settings from the first three improved variations, resulting in an average MPJPE reduction of 7.3%. Baseline 2 employed a three-layer LSTM with speed attenuation adjustment performed by sampling and adding at intervals of 10, 5, and 2. Four fully connected layers were established after the speed incremental learning module, along with a batch normalization layer and a ReLU layer. The sum of five losses was used to update the network weights, and bias was added to all fully connected layers.

The second set of experiments evaluated the influence of individual time interval velocity regulation modules and incremental training on the network's performance. Four variations were compared against Baseline 2: (1) "w/o T=2", "w/o T=5", and "w/o T=10": Eliminating the incremental adjustments with intervals of 2, 5, and 10, respectively. (2) "w/o a": Removing the acceleration correction from Baseline 2. (3) "w/o vf": Eliminating both weighting and acceleration correction from Baseline 2. (4) "0.4 0.6": Prioritizing the most recent two time intervals for weighting.

Compared to Baseline 2, the results showed that: (1) "w/o T=2" reduced the MPJPE of prediction frames greater than 320ms by an average of 6.3%, with decreases ranging from 4.3% to 8.0% across four intervals. (2) "w/o T=5" showed an average reduction of 5.5% compared to Baseline 2, with MPJPE decreasing at shorter intervals (80ms, 160ms, and 320ms) but increasing at intervals over 400ms. (3) "w/o T=10" exhibited a substantial average reduction of 51.0% compared to Baseline 2, with decreases ranging from 6.1% to 81.6% across six intervals. These findings support hypothesis H2, demonstrating that the integration of complementary velocity regulation layers with different time intervals effectively reduces the MPJPE for motion prediction. The velocity adjustment layer with an interval of 10 exhibited the most significant performance enhancement compared to intervals of 2 and 5.

Additionally, the learning rate in "Baseline 2 v2" is reduced by a factor of 5 over Baseline 2, further improving prediction performance. To enhance the network's prediction capabilities, three loss variations were explored: (1) "w/ 6 times loss": Adding the sum of 6 times loss to the original training for weight updates. (2) "w/ 480ms prediction loss": Combining the current loss with the loss of increasing prediction by 80ms. (3) "w/ 600ms prediction loss": Combining the current loss with the loss of increasing prediction by 200ms. An analysis of the results revealed that incorporating these additional losses further enhanced the network's prediction performance, providing further evidence in support of hypothesis H1.

Qualitative comparisons of predictions across several versions (Figures 7 and 8) highlighted the importance of all components in the PMS method. Removing the incremental corrections at intervals of 2, 5, 10, and the acceleration correction led to significant errors up to 1000ms. Specifically, for the "Posing" action, "Baseline 2 v2" exhibited the closest predicted amplitude for body waist bending at 1000ms and the opening amplitude of the two feet, as well as a more accurate prediction of the foot movement at 560ms compared to "Baseline 1". However, the 1000ms prediction of "Baseline 2 v2" significantly deviated from the actual value, indicating the need for further improvements. It is worth noting that the proposed PMS method incorporates a mechanism to mitigate the mean pose problem inherent in recurrent prediction methods. By integrating complementary velocity regulation layers at different time intervals, the PMS method is able to capture and adapt to the temporal dynamics of the human motion, reducing the tendency to converge to a mean pose over longer prediction horizons.

Table 6. Comparison of MPJPE for ablation variants.

| Time(ms) | 80 | 160 | 320 | 400 | 560 | 1000 | Average |
|---|---|---|---|---|---|---|---|
| Baseline 1 | 3.3 | 8.9 | 23.2 | 30.1 | 47.6 | 92.9 | 34.3 |
| FC w/ bias | 3.3 | 8.5 | 22.4 | 29.5 | 46.4 | 91.5 | 33.6 |
| w/ bn+relu | 3.4 | 9.0 | 23.7 | 30.6 | 47.1 | 90.4 | 34.1 |
| w/ 5 times loss | 3.3 | 8.4 | 22.0 | 29.4 | 46.3 | 90.7 | 33.4 |
| w/ 2 5 10 times loss | 3.6 | 9.0 | 24.8 | 32.5 | 50.1 | 95.3 | 35.9 |
| w/ 800ms prediction loss | 4.4 | 10.2 | 24.2 | 31.3 | 48.6 | 93.6 | 35.4 |
| Baseline 2 | 2.8 | 7.4 | 20.2 | 27.7 | 45.3 | 93.3 | 32.8 |
| w/o T=2 | 2.8 | 7.4 | 21.1 | 29.5 | 47.9 | 101.4 | 35.0 |
| w/o T=5 | 3.3 | 8.5 | 22.1 | 28.6 | 46.5 | 99.1 | 34.7 |
| w/o T=10 | 15.2 | 22.1 | 22.0 | 29.5 | 86.0 | 227.5 | 67.0 |
| w/o a | 2.8 | 7.6 | 21.6 | 29.7 | 48.6 | 103.7 | 35.7 |
| w/o vf | 22.1 | 23.2 | 30.1 | 30.2 | 55.1 | 118.2 | 46.5 |
| 0.4 0.6 | 13.9 | 20.2 | 21.6 | 29.9 | 76.5 | 210.9 | 62.2 |
| Baseline 2 v2 | 2.7 | 6.8 | 18.6 | 25.7 | 41.6 | 86.2 | 30.3 |
| w/ 6 times loss | 2.6 | 6.7 | 18.2 | 25.2 | 40.9 | 83.7 | 29.6 |
| w/ 480ms prediction loss | 2.7 | 7.0 | 19.0 | 26.2 | 42.3 | 84.8 | 30.3 |
| w/ 600ms prediction loss | 2.6 | 6.7 | 18.5 | 25.5 | 41.1 | 82.2 | 29.4 |

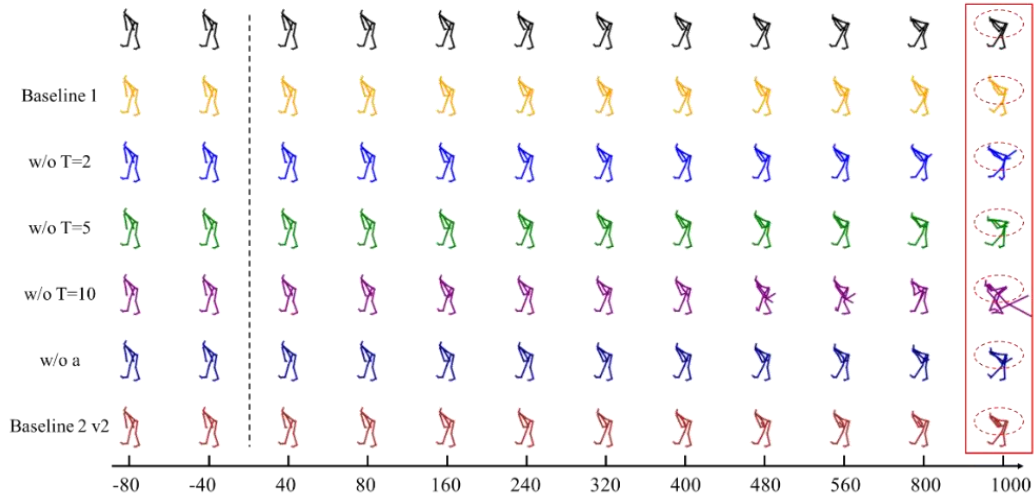

Figure 7. Action of Posing.

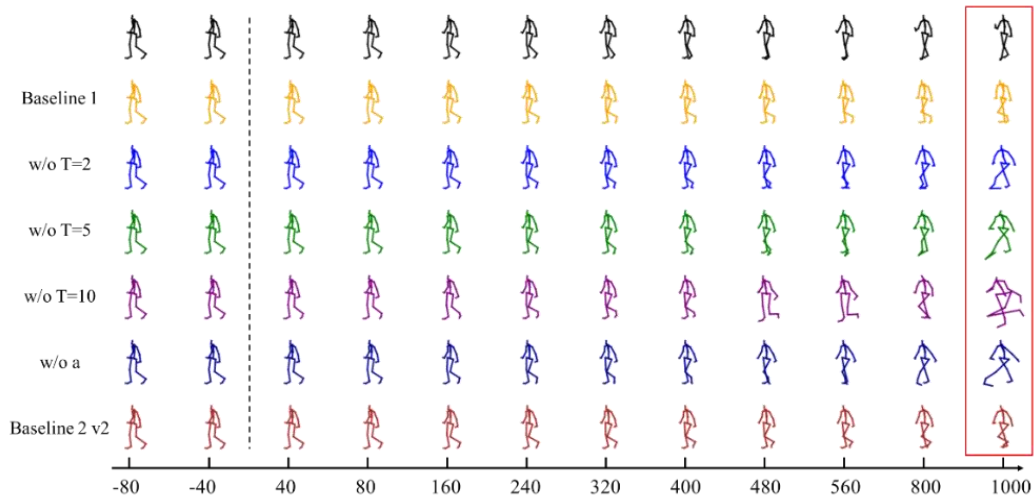

Figure 8. Action of Phoning.

In summary, the experimental results support the hypotheses that network architecture, loss function design, and the integration of complementary velocity regulation layers with different time intervals contribute significantly to the effectiveness of the proposed PMS method in reducing prediction errors and mitigating the mean pose problem for human motion prediction tasks.

# 5 Discussions

While the proposed PMS method demonstrates promising results and outperforms several state-of-the-art approaches, there are some limitations that warrant further investigation and improvement:

1. Prediction of actions with large joint position changes: As observed in the qualitative analysis, the PMS method struggles to accurately predict actions involving substantial changes in joint positions and movements with significant trend alterations. This limitation may arise from the inherent challenges in modeling and capturing the complex dynamics of such motions within the current framework.

2. Generalization across diverse action types: Although the PMS method exhibits improved generalization compared to some existing methods, there is still room for enhancement in terms of consistent performance across a wider range of action types. Certain actions, such as "Basketball" and "Basketball signal" in the CMU dataset, were less effectively learned by the PMS method, suggesting the need for further improvements in the model's ability to generalize to diverse motion patterns.

3. Long-term prediction accuracy: While the PMS method demonstrates improved long-term prediction performance compared to baselines, the prediction errors tend to accumulate and become more significant as the prediction horizon increases, as evidenced by the larger MPJPE values at longer time intervals (e.g., 1000ms).

Future research directions to address these limitations and further enhance the PMS method could involve:

1. Exploring more advanced architectures and techniques capable of better capturing and modeling the complex dynamics of human motion, particularly for actions with large joint position changes and abrupt trend alterations.

2. Incorporating additional data augmentation techniques or transfer learning strategies to improve the model's generalization capabilities across diverse action types and motion patterns.

3. Investigating techniques for mitigating error accumulation in long-term predictions, such as incorporating additional constraints, leveraging hierarchical or multi-scale representations, or employing techniques from sequence-to-sequence learning.

4. Exploring the integration of additional modalities or contextual information (e.g., scene geometry, object interactions) to enhance the model's understanding and prediction of human motion in real-world scenarios.

5. Evaluating the PMS method on larger and more diverse datasets to further validate its performance and identify potential areas for improvement.

By addressing these limitations and exploring new research directions, the PMS method can be further refined and extended, ultimately contributing to more accurate and robust human motion prediction systems with practical applications in various domains.

# 6 Conclusion

This study proposed a novel Parallel Multi-scale Incremental Prediction (PMS) framework for accurate human motion forecasting. The primary contributions are threefold:1) A multi-scale computation module capturing motion dynamics across multiple time scales by modeling velocity and acceleration differences in parallel branches. 2) A fusion module synthesizing multi-scale dynamics, enabling iterative differential learning and correction of motion predictions. 3) An optimized full-time loss function incorporating long-range context to better capture short-term and long-term dependencies. Extensive experiments validated PMS's efficacy, achieving state-of-the-art performance with 16.3%-64.2% accuracy improvements over previous methods by enhancing biomechanical consistency, continuity, and long-term stability.

The significance and potential impact of this work are broader than the technical contributions alone. By achieving more accurate and stable long-term motion prediction, PMS paves the way for safer, seamless human-robot collaboration across many realms including manufacturing, healthcare services, home assistance care, and beyond. The abilities to forecast complex human behaviors and preemptively adapt could revolutionize how machines interact with and aid people in diverse environments. We encourage further exploration into alternative representations, attentions mechanisms, and contextual fusion techniques to handle even more erratic, rapidly changing motions over long horizons. There remain open challenges, but the advancements made here impart momentum and inspiration to enhance prediction capabilities—not just marginal gains in accuracy metrics, but toward unraveling the intricacies of dynamic motor coordination through modeling, prediction algorithmic innovations. The future promises great progress at this exciting intersection bridging data sciences, human movement biomechanics, and interactive robotics.